\documentclass[runningheads,a4paper]{llncs}

\usepackage{amssymb}
\usepackage{graphicx}
\usepackage{epstopdf}
\usepackage{subfigure}
\usepackage{float}
\usepackage{color}
\usepackage{makeidx}
\usepackage{amsmath}
\usepackage{multirow}

\usepackage{url}
\usepackage[table]{xcolor}
\usepackage{array}
\newcolumntype{C}[1]{>{\centering\let\newline\\\arraybackslash\hspace{0pt}}m{#1}}
\newcolumntype{R}[1]{>{\raggedleft\let\newline\\\arraybackslash\hspace{0pt}}m{#1}}

\newcommand{\keywords}[1]{\par\addvspace\baselineskip
\noindent\keywordname\enspace\ignorespaces#1}

\begin{document}

\mainmatter  % start of an individual contribution

% \title{Modeling the Event-driven  Temporal Evolution of Knowledge Graphs}
% \title{A Prediction Model  for the  Co-Evolution of Event  and Knowledge Graphs}
\title{Predicting  the  Co-Evolution of Event  and Knowledge Graphs}
% Modeling the Event-driven  Temporal Evolution of Knowledge Graphs}

 \titlerunning{Predicting  the  Co-Evolution of Event  and Knowledge Graphs}

\author{Crist\'{o}bal  Esteban$^{1,2}$  \and Volker Tresp$^{1,2}$  \and Yinchong Yang$^{2}$ \and Stephan Baier$^{2}$ \and Denis Krompa\ss$^{1}$}
\authorrunning{C. Esteban \and  V. Tresp  \and  Y. Yang \and  S. Baier \and D.   Krompa\ss}
\institute{
Siemens AG, Corporate Technology, Munich, Germany\\
\email{\{Cristobal.Esteban\}\{Volker.Tresp\}\{Denis.Krompass\}@siemens.com}\\
%\email{Volker.Tresp@siemens.com}\\
%\email{Denis.Krompass@siemens.com}\\
Ludwig Maximilian University of Munich, Germany\\
 \email{\{Stephan.Baier\}\{Yinchong.Yang\}@campus.lmu.de}
}

\maketitle
\begin{abstract}

Embedding learning, a.k.a. representation learning,  has been shown to be able to model large-scale semantic knowledge graphs. A  key concept is a mapping of the knowledge graph to  a tensor representation whose entries are predicted by models using latent representations of generalized entities.
Knowledge graphs are typically treated as static: A knowledge graph  grows more links when more facts become available but the ground truth values associated with  links is considered time invariant.
In this paper we address the issue of knowledge graphs where triple states depend on time.
We assume that changes in the knowledge graph always arrive in form of events, in the sense that the events are the gateway to the knowledge graph.
We train an event prediction model which  uses both knowledge graph background information  and  information on recent events.    By predicting future events, we also predict likely changes in the knowledge graph and thus obtain a model for the evolution of the knowledge graph as well.
 Our experiments demonstrate that  our approach performs well in  a clinical application, a recommendation engine and a sensor network application.

\end{abstract}

%%%%%%%%%%%%%%%%%%%%%%%%%%%%%%%%%%%%%%%%%%%%%%%%%%%%%%%%%%%%%%%%%%%%%%%%%%%%%%%%%%%%%%
% INTRODUCTION
%%%%%%%%%%%%%%%%%%%%%%%%%%%%%%%%%%%%%%%%%%%%%%%%%%%%%%%%%%%%%%%%%%%%%%%%%%%%%%%%%%%%%%

\section{Introduction}
\label{sec:introduction}

In previous publications it was shown how  a triple knowledge graph (KG)  can be represented as a  multiway array (tensor) and how a statistical  model can be formed  by  deriving latent representations of generalized entities. Successful models are,  e.g., RESCAL~\cite{nickel_three-way_2011}, Translational Embeddings Models~\cite{bordes_learning_2011},  Neural Tensor Models~\cite{socher_reasoning_2013} and the  multiway neural networks as used in~\cite{dong_knowledge_2014}.
In these publications   KGs were treated as static: A KG  grew more links when more facts became available but the ground truth value associated with a link was considered time invariant.
In this paper we address the issue of KGs where triple states depend on time. In the simplest case this might consider simple facts like ``Obama is president of the United States'', but  only from 2008-2016. Another example is a patient whose status changes from sick to healthy or vice versa. Most popular KGs like Yago~\cite{suchanek_yago:_2007}, DBpedia~\cite{auer_dbpedia:_2007}
Freebase~\cite{bollacker_freebase:_2008} and the Google Knowledge Graph~\cite{singhal_introducing_2012} have means to store temporal information.

Without loss of generality, we assume that changes in the KG always arrive in form of events, in the sense that the events are the gateway to the KG. For a given time step, events are described by a typically very sparse event  triple graph, which contains facts that  change some of the triples in the KG, e.g., from \textit{True} to \textit{False} and vice versa. KG triples which do not appear in the event graph are assumed unchanged.

An example might be the statement that a patient has a new diagnosis of diabetes, which is an information that first appears as an event in the event  graph but is then also transmitted to the KG.    Other events might be a prescription of a medication to lower the cholesterol level,  the decision to measure the cholesterol level  and the measurement result  of the cholesterol level; so events can be, e.g.,  actions,  decisions and measurements. In a similar way as the KG is represented as a KG tensor, the event graphs for all time steps can be represented as an event tensor. Statistical models for both the KG tensor and the event tensor can be derived based on  latent representations derived from the tensor contents.

Although the event tensor has a representation for  time, it is by itself not a prediction model.  Thus, we train a separate prediction model which estimates future events based on the latent representations of previous events in the event tensor and the latent representations of the involved generalized entities in the KG tensor. In this way,  a prediction can, e.g.,  use both background information describing the status of a patient and can consider recent events.    Since some future events will be absorbed into the KG, by predicting future events, we also predict likely changes in the KG and thus obtain a model for the evolution of the KG as well.

The paper is structured as follows. The next section discusses related work.
In Section~\ref{sec:tensors} we review statistical KG models based on latent representations of the involved generalized entities.
In  Section~\ref{sec:events} we discuss the event tensor and its latent representations.
 In Section~\ref{sec:markov}  we demonstrate how future events can be estimated using a prediction model that uses  latent representations of the KG model and the event model as inputs.
In the applications of this paper in Section~\ref{sec:exper}, we consider patients and their changing states,  users and their changing movie preferences and weather stations and their  changing signal statistics.
In the latter  we show how ---in addition to event data--- sensory data can be modeled.
  Section~\ref{sec:concl} contains our conclusions and discusses extensions.

\section{Related Work}
\label{sec:related}

There is a wide range of papers on the application of  data mining and machine learning to KGs. Data mining attempts to find interesting KG patterns~\cite{berendt2002towards,rettinger2012mining,paulheim2014data}. Some machine learning approaches attempt to extract close-to deterministic dependencies and ontological constructs~\cite{maedche2001ontology,fanizzi2008dl,lehmann2009dl}. The paper here focuses on \emph{statistical} machine learning in  KG where representation learning has been proven to be very successful.

There is considerable prior work on  the application of tensor models to temporal data, e.g., EEG data,  and  overviews can be found in~\cite{kolda_tensor_2009} and~\cite{morup}. In that work,  prediction is typically not in focus, but instead one attempts to understand essential underlying temporal processes by analysing the derived latent representations.

Some models consider a temporal parameter drift. Examples are  the BPTF~\cite{DBLP:conf/sdm/XiongCHSC10}, and
~\cite{DBLP:journals/tkdd/DunlavyKA11}. Our model has a  more expressive dynamic by explicitly considering recent histories. Markov properties in tensor models were considered in~\cite{RendleFS10,rendle_factorization_2010}. In that work quadratic interactions between latent representations were considered. The approach described here is  more general  and also considers multiway neural networks as flexible function approximators.

Our approach can also be related to the
  neural probabilistic language
model~\cite{bengio1}, which coined the term \emph{representation learning}. It can be considered an event model  where the occurrence of a word is predicted based on most recent observed words using a neural network model with word representations as inputs. In our approach we consider that several events might be observed at a time instance and we consider a richer family of latent factor representations.

There is considerable recent work on dynamic graphs~\cite{DBLP:conf/kdd/LeskovecKF05,DBLP:journals/sigkdd/OMadadhainHS05,DBLP:conf/kdd/SunFPY07,DBLP:conf/wsdm/RossiGNH13} with a strong focus on the Web graph and social graphs. That work is not immediately applicable to KGs but we plan to explore potential links as part of our future work.

\section{The Knowledge Graph  Model}
\label{sec:tensors}

With the advent of  the Semantic Web~\cite{berners-lee_semantic_2001}, Linked Open Data~\cite{berners-lee_linked_2006}, Knowledge Graphs (KGs)~\cite{suchanek_yago:_2007,auer_dbpedia:_2007,bollacker_freebase:_2008,singhal_introducing_2012},
triple-oriented knowledge representations have gained in popularity.
Here we consider a slight extension to    the subject-predicate-object triple form by adding the value
($e_s, e_p, e_o$; \textit{Value}) where \textit{Value} is a function of $s, p, o$ and can be the truth value of the triple or it can be  a measurement. Thus \textit{(Jack, likes, Mary; True)} states that Jack likes Mary,
and \textit{(Jack, hasBloodTest, Cholesterol; 160)} would indicate a particular blood cholesterol level for Jack. Note that $e_s$ and $e_o$ represent the entities for subject index $s$ and object index $o$. To simplify notation we also consider $e_p$ to be a generalized  entity associated with predicate type with index $p$.

A machine learning approach to inductive inference in KGs is based on the factor analysis of its adjacency tensor  $\underline{X}$
where the  tensor element $x_{s, p, o}$  is the associated \textit{Value} of the triple ($e_s, e_p, e_o$). Here $s = 1, \ldots, S$, $p = 1, \ldots, P$, and $o = 1, \ldots, O$.
 One can also define a second tensor $\underline{\Theta}^{KG}$ with the same dimensions as $\underline{X}$.  It contains the natural parameters of the model and the connection to $\underline X$.  In the binary case one can use a Bernoulli likelihood with
 $P(x_{s, p, o} | \theta^{KG}_{s, p, o} )    \sim
\textrm{sig}(\theta^{KG}_{s, p, o})$, where
$\textrm{sig}(arg) = 1/(1+\exp(-arg))$ is the logistic function.  If $x_{s, p, o}$ is a real number than we can use a Gaussian distribution with  $P(x_{s, p, o} | \theta^{KG}_{s, p, o} ) \sim  {\mathcal{N}}(\theta^{KG}_{s, p, o}, \sigma^2)$.

In representation learning, one assigns an $r$-dimensional latent vector to the entity $e$ denoted by
$
\mathbf{a}_e  = (a_{e, 0}, a_{e, 1},  \ldots, a_{e, r})^T
$.
We then model using one function
 \[
 \theta^{KG}_{s, p, o} = f^{KG}(\mathbf{a}_{e_s}, \mathbf{a}_{e_p}, \mathbf{a}_{e_o})
 \]
or,  using one function for each predicate,
 \[
 \theta^{KG}_{s, p, o} = f^{KG}_p(\mathbf{a}_{e_s}, \mathbf{a}_{e_o}) .
 \]

For example, the   RESCAL model~\cite{nickel_three-way_2011} is
 \[
\theta^{KG}_{s, p, o} = \sum_{k=1}^r \sum_{l=1}^r R_{p, k, l} a_{e_s,k}  a_{e_o,l}  ,
\]
where $R \in \mathbb{R}^{P \times r \times r}$ is the core tensor.
In the multiway neural network model~\cite{dong_knowledge_2014} one uses
 \[
\theta^{KG}_{s, p, o} = \textrm{NN} (\mathbf{a}_{e_s}, \mathbf{a}_{e_p},  \mathbf{a}_{e_o})
\]
where $\textrm{NN}$ stands for a neural network and where the inputs are concatenated.  These approaches have been used very successfully to model large KGs, such as the Yago KG, the DBpedia KG and parts of the Google KG.  It has been shown experimentally that models using latent factors perform well in these high-dimensional and highly sparse domains. For a recent review, please consult~\cite{nickel2015}.

We also consider an alternative representation.
The idea is that  the latent vector stands for the tensor entries associated with the corresponding entity.  As an example, $\mathbf{a}_{e_{s}}$ is the latent representation for all values associated with entity $e_s$, i.e., $x_{s, :, :}$. \footnote{If an entity can also appear as an object ($o: e_o = e_s$), we need to include
$x_{:, :, o}$. }
It is  then convenient to
 assume that one can calculate a so-called $M$-map of the form
 \begin{equation}\label{eq:map}
   \mathbf{a}_{e_{s}} = M^{\textit{subject}}  x_{s, :, :}  .
 \end{equation}
Here $M^{\textit{subject}} \in \mathbb{R}^{r \times (P O)}$ is a mapping matrix to be learned and $x_{s, :, :}$ is a column vector of size $PO$.\footnote{The $M$ matrices are dense but one dimension is small ($r$), so in our settings we did not run into storage problems. Initial experiments indicate that  random projections can be used in case that computer memory becomes a limitation.} For multilinear models  it can be shown that such a representation is always possible; for other models this is a constraint on the latent factor representation. The advantage now is that the latent representations of an entity can be calculated in one simple vector matrix product, even for new entities not considered in training.
 We can define similar maps for all latent factors. For a given latent representation we can either  learn the latent factors directly, or we learn an  $M$-matrix.

 The latent factors, the $M$-matrices, and the parameters in the functions  can be trained with  penalized log-likelihood cost functions described in the Appendix.

\section{The Event Model}
\label{sec:events}

 Without loss of generality, we assume that changes in the KG always arrive in form of events, in the sense that the events are the gateway to the KG. For a given time step, events are described by a typically very sparse event  triple graph, which contains facts that  change some of the triples in the KG, e.g., from \textit{True} to \textit{False} and vice versa. KG triples which do not appear in the event graph are assumed unchanged.

Events  might be,  e.g.,
\textit{do a cholesterol measurement},  the event \textit{cholesterol measurement}, which specifies the value or the order \textit{take cholesterol lowering medicine}, which determines that a particular medication is prescribed followed by dosage information.

At each time step events form triples which form a sparse triple graph and which specifies which facts become available. % The event graph is typically very sparse and some triples in the event graph imply changes in the KG.
The event tensor is a  four-way tensor $\underline Z$ with $(e_s, e_p, e_o, e_t; \textit{Value})$ and tensor elements $z_{s, p, o, t}$.  We have introduced the generalized entity $e_t$ to represent time. Note that the characteristics of the KG tensor and the event tensor are quite different.   $\underline X$ is sparse and entries rarely change with time. $\underline  Z$ is even sparser and nonzero entries typically ``appear'' more random.    We model
\[
\theta^{event}_{s, p, o} = f^{event} (\mathbf{a}_{e_s}, \mathbf{a}_{e_p},  \mathbf{a}_{e_o}, \mathbf{a}_{e_t})  .
\]
Here,  $\mathbf{a}_{e_t}$ is the latent representation of the generalized entity $e_t$.

Alternatively, we consider  a personalized representation of the form
\[
\theta^{\textit{pers-event}, s=i}_{p, o} = f^{\textit{pers-event}} (\mathbf{a}_{e_p},  \mathbf{a}_{e_o}, \mathbf{a}_{e_s=i, t})  .
\]
Here, we have introduced the generalized entity $e_{s, t}$ for a subject $s=i$ at time $t$ which stands for all events of entity $s=i$ at time $t$.

Since representations involving time need to be calculated online, we use M-maps of the form
 \[
   \mathbf{a}_{e_{s, t}} =  M^{\textit{subject, time}}  z_{s, :, :, t}
 \]

The cost functions  are again described in the Appendix.

\section{The Prediction Model}
\label{sec:markov}

\subsection{Predicting Events}

Note that  both the KG-tensor  and the event tensor can only model information that was observed until
 time $t$ but it would not be easy to derive predictions for future events, which would be of interest, e.g., for decision support.   The key idea of the paper is that events are predicted using both latent representations of the KG and latent representations describing recently observed events.

In the prediction model we estimate future entries in  the event tensor ${\underline Z}$.  The general form is
\[
\theta^{\emph{predict}}_{s, p, o, t} =
f^{\emph{predict}}(\emph{args}) \;\;\; \textrm{or }\;\;\; \theta^{\emph{predict}}_{s, p, o, t} =
f^{\emph{predict}}_{p, o}(\emph{args})
\]
where the first version uses a single  function and the latter  uses a  different function for each $(p,o)$-pair.\footnote{The different functions can be realized by the multiple outputs of a neural network.}  Here, $\emph{args}$ is from the sets of latent representations from the KG tensor and  the event tensor.

An example of a prediction model is
\[
\theta^{\textit{predict}}_{s, p, o, t}  =
f^{\textit{predict}}_{p, o}( \mathbf{a}_{e_{s}},  \mathbf{a}_{e_{s, t}},   \mathbf{a}_{e_{s, t-1}}, \ldots, \mathbf{a}_{e_{s, t-T}}) .
\]
where the prediction is based on the latent representations of subject, object and predicate from the KG-tensor and of the time-specific representations from the event  tensor.

Let's consider  an example. Let $(e_{s}, e_p, e_o, e_t; \textit{Value})$ stand for
(\textit{Patient, prescription, CholesterolMedication, Time; True}).
Here,  $\mathbf{a}_{e_{s}}$   is the  profile of the patient, calculated from the KG model. Being constant,  $\mathbf{a}_{e_{s}}$ assumes the role of  parameters in the prediction model.
  $\mathbf{a}_{e_{s, t}}$ describes all that \emph{so far} has happened to the patient at the same instance in  time $t$  (e.g., on the same day).
 $\mathbf{a}_{e_{s, t-1}}$ describes all that happened to the patient at the last instance in time and so on.

We model the functions  by a multiway neural network with weight parameters $W$ exploiting the great modeling flexibility of neural networks. The cost function for the prediction model is
\begin{equation}\label{eq:cost}
\textrm{cost}^{\textit{predict}} =
-  \sum_{z_{s, p, o, t} \in \underline{{Z}}}  \log P(z_{s, p, o, t} | \theta^{\textit{predict}}_{s, p, o, t}
(A, M, W))
\end{equation}
\[+ \lambda_{A} \| A \|_F^2 + \lambda_{W} \| W \|_F^2 + \lambda_{M} \| M \|_F^2  .
\]
$A$ stands for the parameters in latent representation and $M$  stands for the parameters in the $M$-matrices. For a generalized entity for which we use an $M$-matrix, we penalize the entries in the $M$-matrix; for a generalized entity for which we directly estimate the latent representation we penalize the entries in the corresponding latent terms in $A$.
Here,  $\| \cdot \|_F$ is the Frobenius norm and   $\lambda_{A}\ge 0$, $\lambda_{M}\ge 0$ and  $\lambda_{W}\ge 0$   are regularization parameters.

\subsection{Predicting Changes in the KG}

\begin{figure}[ht]
	\centering
	\includegraphics[width=0.6\columnwidth]{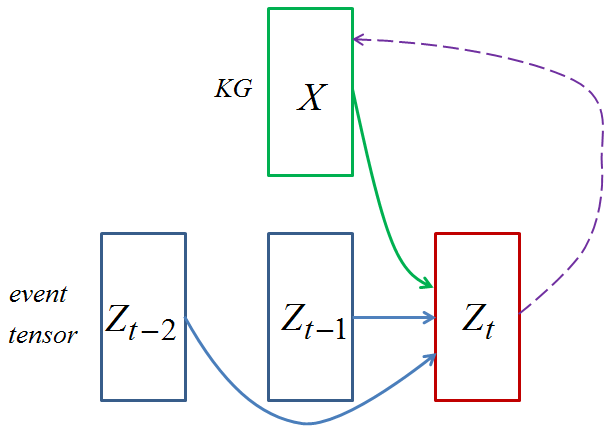} %ESWC-Architectures/Folie14.png}
	\caption{%
		The  figure shows an example where  the event tensor is predicted from the representations of the  events in the last two time steps and from the KG representation. The dotted line indicate the transfer of observed events into the KG.
	}
	\label{fig:TMs}
\end{figure}

In our model, each change in the status of the KG is communicated via events. Thus each change in the KG first appears in the event tensor and predictions of events also implies predictions in the KG.    The events that change the KG status are transferred into the KG and the latent representations of the KG, i.e., $\mathbf{a}_{e_{s}}, \mathbf{a}_{e_{p}}, \mathbf{a}_{e_{o}}$,  are  re-estimated regularly (Figure~\ref{fig:TMs}).

\subsection{More Cost Functions}

Associated with each tensor model and prediction model, there  is a cost function (see Appendix). In our experiments we obtained best results, when we used the cost function of the task we are trying to solve.  In the most relevant prediction task we thus use the cost function in Equation~\ref{eq:cost}. On the other hand, we obtained faster convergence for the prediction model if we initialize latent representations  based on the KG model.

\section{Experiments}
\label{sec:exper}

\begin{figure}[ht]
	\centering
	\includegraphics[width=0.8\columnwidth]{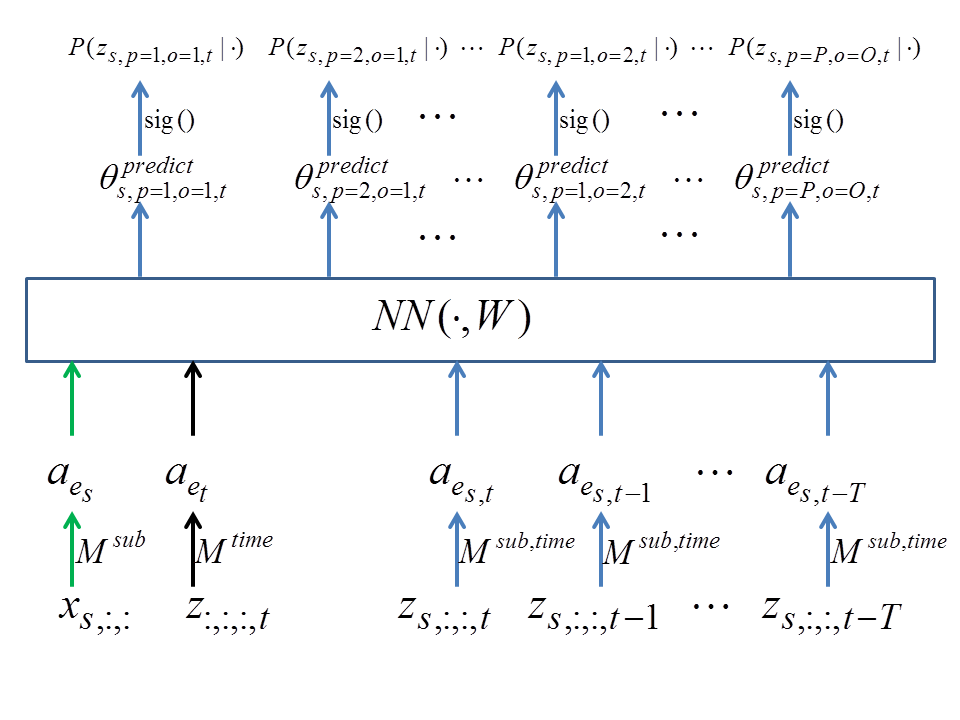}

	\caption{%
	The prediction  model for the clinical data.
	}
	\label{fig:archs1}
\end{figure}

\begin{figure}[ht]
	\centering
	\includegraphics[width=0.8\columnwidth]{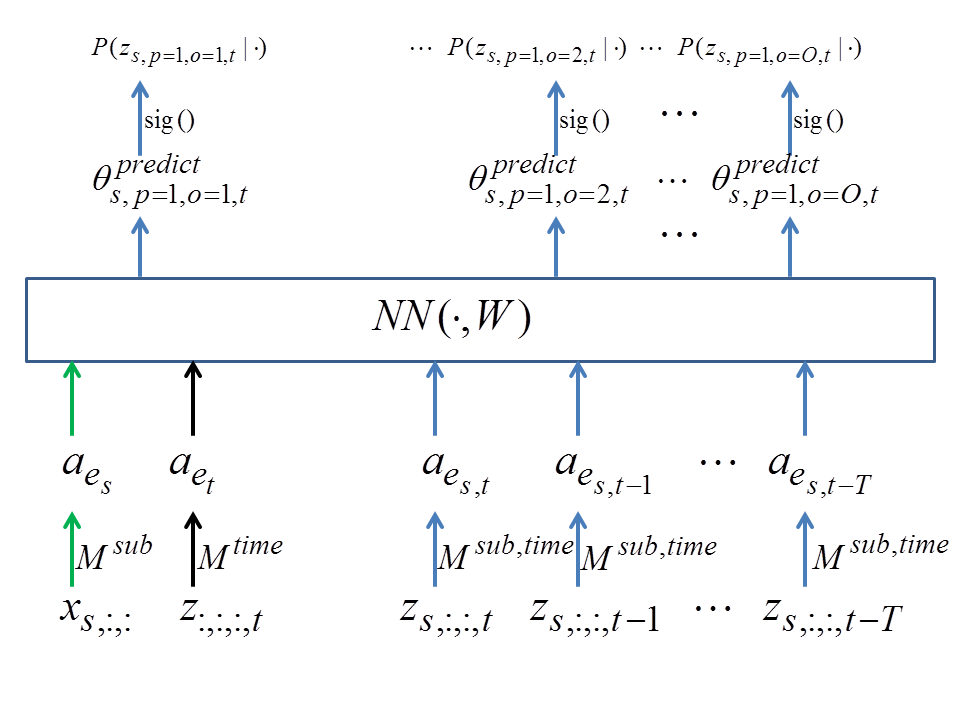}

	\caption{%
		 The prediction model for the recommendation data.
	}
	\label{fig:archs2}
\end{figure}

\begin{figure}[ht]
	\centering
	\includegraphics[width=0.8\columnwidth]{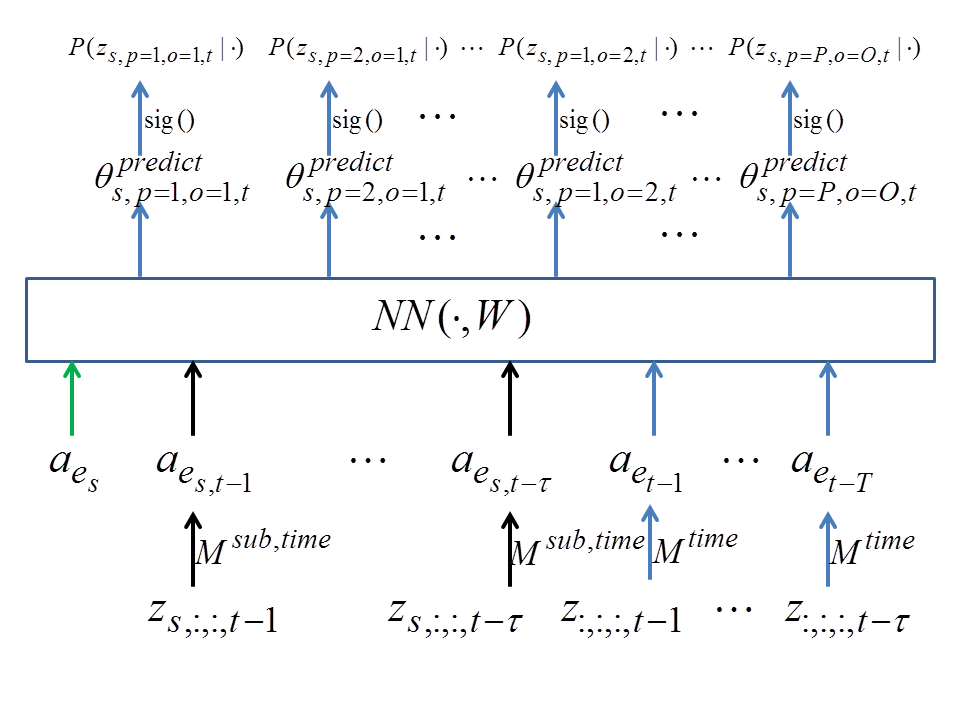}

	\caption{%
		 The prediction model for the sensor data. $\textbf{a}_{e_s}$ is directly estimated without using an $M$-mapping.
	}
	\label{fig:archs3}
\end{figure}

\subsection{Modeling Clinical Data}
\label{sec:patient}

\begin{table*}[!htbp]
\caption[Table caption text]{Scores for next visit predictions. AUPRC stands for Area Under Precision-Recall Curve. AUROC stands for Area Under ROC Curve. \textit{ET} stands for our proposed model that uses only past event information but no information from the KG. }
\centering
\begin{tabular}{lcccc}
\hline\rule{0pt}{12pt}
                     & {\hskip 0.05cm}  AUPRC  {\hskip 0.05cm}  & AUROC   {\hskip 0.05cm}   & Time (hours)      \\
\hline \\ [-1.8ex]
\emph{ET}       & {\hskip 0.05cm}  0.574 $\pm$ 0.0014  {\hskip 0.05cm}  & 0.977 $\pm$ 0.0001  {\hskip 0.05cm}   & 6.11      \\
Logistic Regression  & {\hskip 0.05cm}  0.554 $\pm$ 0.0020  {\hskip 0.05cm}  & 0.970 $\pm$ 0.0005  {\hskip 0.05cm}   & 4.31            \\
KNN				     & {\hskip 0.05cm}  0.482 $\pm$ 0.0012   {\hskip 0.05cm}  & 0.951 $\pm$ 0.0002  {\hskip 0.05cm}   & 17.74            \\
Naive Bayes          & {\hskip 0.05cm}  0.432 $\pm$ 0.0019  {\hskip 0.05cm}  & 0.843 $\pm$ 0.0015  {\hskip 0.05cm}   & 39.1            \\
Constant predictions & {\hskip 0.05cm}  0.350 $\pm$ 0.0011  {\hskip 0.05cm}  & 0.964 $\pm$ 0.0001  {\hskip 0.05cm}   & 0.001            \\
Random				 &  {\hskip 0.05cm}  0.011 $\pm$ 0.0001  {\hskip 0.05cm}  & 0.5    {\hskip 0.05cm}   & -            \\

\hline\rule{0pt}{12pt}
\end{tabular}
\label{table:table_exp1}
\end{table*}

The study is based on  a large data set collected from  patients that
suffered from kidney failure. The data was collected  in
the Charit\'{e} hospital  in Berlin  and it is the largest data collection  of its kind  in Europe. Once the kidney has failed, patients face
a lifelong treatment and periodic visits to the clinic for the rest of their
lives.
After the
transplant has been performed, the patient receives immunosuppressive therapy
to avoid the rejection of the transplanted kidney. The patient must be controlled
periodically to check the status of the kidney, adjust the treatment and take
care of associated diseases, such as those that arise due to the
immunosuppressive therapy.
 The dataset contains every event that happened to each
patient concerning the kidney failure and all its associated events:
prescribed medications, hospitalizations, diagnoses, laboratory tests, etc. ~\cite{Lindemann2007,Schroeter2010}.  The database started being recorded
more than 30 years ago and it is composed of dozens of tables with  more than 4000 patients that
underwent a renal transplant or are waiting for it. For example, the
database contains more than 1200 medications that have been prescribed more
than 250000 times, and the results of more than 450000 laboratory analyses.

 This is particularly important for the estimation of drug-drug interactions (DDI) and
adverse drug reactions (ADR) in patients after renal transplant.

We  work with a subset of the variables available in the dataset. Specifically, we  model medication prescriptions, ordered lab tests and lab test results.
We transformed the tables into an event oriented representation where the subject is the patient and where time is a patient visit. We encoded the lab results in a binary format representing normal, high, and low values of a lab measurement, thus \textit{Value} is always binary.

The prediction model  is
\[
\theta^{predict}_{s, p, o, t} =
f^{predict}_{p, o}(
{\mathbf{a}}_{e_{s}},
{\mathbf{a}}_{e_t},
% {\mathbf{a}}_{e_{p,o,t}},
{\mathbf{a}}_{e_{s,t}},
{\mathbf{a}}_{e_{s, t-1}}, \ldots,
{\mathbf{a}}_{e_{s, t-T}}   ) .
\]
Note that we have a separate function  for each $(p,o)$-pair.
$ {\mathbf{a}}_{e_{s}}$ are patient properties as described in the KG.
$\mathbf{a}_{e_{s, t}}$ represents  all events known that happened at visit $t$ for the patient (e.g., the same visit for which we want to make a prediction).
$\mathbf{a}_{e_{s, t-1}}$ represents all events for the patient  at the last visit, etc.
${\mathbf{a}}_{e_t}$ stands for the latent representation of all  events at visit $t$ for all patients and can model if events are explicitly dependent on the time since the transplant.
  Regarding the input  window, we empirically found that  $T=6$ is optimal.
 The architecture is shown in Figure~\ref{fig:archs1}.

 The first experiment consisted of predicting the events that will happen to patients in their  next visit to the clinic given the events that were observed in the patients' previous visits to the clinic (i.e. by using the events that occurred to the patient from $\mathbf{a}_{e_{s, t}}$ until $\mathbf{a}_{e_{s, t-6}}$). The experiment was performed 10 times with different random splits of the patients. Thus we truly predict performance on patients which were not considered in training!
 Table~\ref{table:table_exp1} shows how our proposed model outperforms the baseline models.
 The ``constant predictor'' always predicts for each event the occurrence
rate of such event (thus the most common event is given
the highest probability of happening, followed by the second
most common event, and so on).
Note that we are particularly interested in the Area Under Precision-Recall Curve score due to the high sparsity of the data and our interest in predicting events that will actually happen, as opposed to the task of predicting which events will not be observed. In the last column of Table I we also report the time that it took to train for each model with the best set of hyperparameters in the first random split.

 Next we repeat the experiment including the KG-representation of the patient, which contains static variables of the patient such as blood type and gender,  i.e.,  $ {\mathbf{a}}_{e_{s}}$, and also used  ${\mathbf{a}}_{e_t}$.
  Table \ref{table:table_exp4} shows the improvement brought by the inclusion of the KG representation. The last row in Table \ref{table:table_exp4} shows the result of making the predictions just with the KG representation of the patient (i.e. without the  past event information), demonstrating clearly that information on past events is necessary to achieve best performance.

\begin{table}[ht]
\caption[Table caption text]{Scores for full visit predictions with and without the information in the KG. AUPRC stands for Area Under Precision-Recall Curve. AUROC stands for Area Under ROC Curve. \textit{ET+KG} stands for our proposed model that uses past event information and information from the KG. \emph{ET} only uses past event data and \emph{KG} only uses KG data.}
\centering
\begin{tabular}{lcccc}
\hline\rule{0pt}{12pt}
                     & {\hskip 0.3cm}  AUPRC  {\hskip 0.3cm}  & AUROC   {\hskip 0.3cm}   \\
\hline \\ [-1.8ex]
\textit{ET+KG}      & {\hskip 0.3cm}  0.586 $\pm$ 0.0010  {\hskip 0.3cm}  & 0.979 $\pm$ 0.0001  {\hskip 0.3cm}  \\
\textit{ET}      & {\hskip 0.3cm}  0.574 $\pm$ 0.0014  {\hskip 0.3cm}  & 0.977 $\pm$ 0.0001  {\hskip 0.3cm}  \\
\textit{KG}       & {\hskip 0.3cm}  0.487 $\pm$ 0.0016  {\hskip 0.3cm}  & 0.974 $\pm$ 0.0002  {\hskip 0.3cm}  \\
\hline\rule{0pt}{12pt}
\end{tabular}
\label{table:table_exp4}
\end{table}

\subsection{Recommendation Engines}
\label{sec:recommendation}

We used data from the MovieLens project with 943 users and 1682 movies.\footnote{http://grouplens.org/datasets/movielens/}
In the KG tensor we considered the triples
 \textit{(User, rates, Movie; Rating)}. For the event tensor, we considered
 the quadruples
\textit{(User, watches, Movie, Time; Watched)} and \textit{(User, rates, Movie, Time; Rating)}.
Here,  $\textit{Rating} \in \{1, \ldots, 5 \}$ is the score the user assigned to the movie and $\textit{Watched} \in \{0, 1\}$ indicates if the movie was watched and rated at time $t$.
 \textit{Time} is the calendar week of the rating event.
We define our training data to be 78176 events in the first 24 calender weeks and the test data to be 2664 events in the last 7 weeks. Note that in both datasets there are only 738 users since the remaining 205 users watched and rated their movies only in the test set.

It turned out that the movie ratings did not show   dependencies on past events, so they could be predicted from the KG model alone with
\[
\theta^{predict}_{s, \textit{rates},  o, t} =
f^{predict}_{\textit{rates}, o}( {\textbf{a}}_{e_{s}})  .
\]
We obtained best results by modeling  the function with a neural network with 1682 outputs (one for each movie).
The user specific data  was  centered w.r.t. to their average and a numerical 0 would stand for a neutral \textit{rating}. We obtain an RMSE score of 0.90 $\pm$ 0.002 which is competitive with the best reported score of 0.89 on this data set~\cite{rendle_factorization_2010}. But note that we predicted \textit{future ratings} which is more difficult than predicting randomly chosen test ratings, as done in the other studies. Since we predict ordinal ratings, we used a Gaussian likelihood model.

Of more interest in this paper is to predict if a user will decide to watch a movie at the next time step. We used  a prediction model with
\[
\theta^{predict}_{s, \textit{watches}, o , t } =
\]
\[
f^{predict}_{\textit{watches}, o}(
 {\mathbf{a}}_{e_{s}},
 {\mathbf{a}}_{e_t},
{\mathbf{a}}_{e_{\textit{s, t}}}, {\mathbf{a}}_{e_{s, t-1}}, \ldots,   {\mathbf{a}}_{e_{s, t-T}}
)  .
\]
Here,  $\mathbf{a}_{e_{s}}$ stands for the profile of the user as represented in the KG.
$\mathbf{a}_{e_t}$  stands for the latent representation of all events at time $t$ and can model seasonal preferences for movies.
$\mathbf{a}_{e_{s, t}}$ stands for the latent representation of all movies that the user  watched at time $t$.  The architecture is shown in Figure~\ref{fig:archs2}.
When training with only the prediction cost function  we observe an AUROC an score of 0.728 $\pm$ 0.001.
We then explored sharing of statistical  strength by  optimizing jointly  the $M$-matrices using all three cost functions  $\textrm{cost}^\textit{KG}, \textrm{cost}^\textit{event}$ and $\textrm{cost}^\textit{predict}$ and obtained  a significant improvement with  an AUROC score of 0.776 $\pm$ 0.002.

For comparison, we considered  a pure  KG-model
and achieved an AUROC score of 0.756 $\pm$ 0.007. Thus  the information on past events leads to a   small (but significant) improvement.

\subsection{Sensor Networks}
\label{sec:sensor}

\begin{table}[ht]
	\caption[Table caption text]{Mean Squared Error scores for predicting multivariate sensor data 20 time steps ahead.}
	\centering
	\begin{tabular}{lccc}
		\hline\rule{0pt}{12pt}
		Model & {\hskip 0.3cm}  MSE  {\hskip 0.3cm}   \\
		\hline \\ [-1.8ex]
		Pred1       & {\hskip 0.3cm}  0.135 $\pm$ 0.0002  {\hskip 0.3cm}   \\
		Pred2      & {\hskip 0.3cm}  0.139 $\pm$  0.0002 {\hskip 0.3cm}   \\
		Pred3     & {\hskip 0.3cm}  0.137 $\pm$  0.0002  {\hskip 0.3cm}   \\
		Feedforward Neural Network     & {\hskip 0.3cm}   0.140 $\pm$  0.0002  {\hskip 0.3cm}   \\
		Linear Regression     & {\hskip 0.3cm}   0.141 $\pm$  0.0001  {\hskip 0.3cm}   \\
		Last Observed Value    & {\hskip 0.3cm}  0.170   {\hskip 0.3cm}   \\
		\hline\rule{0pt}{12pt}
	\end{tabular}
	\label{table:table_sensor_results}
\end{table}

In our third experiment we wanted to explore if our approach is also applicable to data from sensor networks.
The main difference is now that the event tensor becomes a sensor tensor with subsymbolic measurements at all sensors at all times.

Important research issues for wind energy systems concern  the accurate wind profile prediction, as it plays an important role in planning and designing of wind farms. Due to the complex intersections among large-scale geometrical parameters such as surface conditions, pressure, temperature, wind speed and wind direction, wind forecasting has been considered a very challenging task. In our analysis we used data from the Automated Surface Observing System (ASOS) units that are operated and controlled cooperatively in the United States by the NWS, FAA and DOD\footnote{http://www.nws.noaa.gov/asos/}. We downloaded the data from the Iowa Environmental Mesonet (IEM)\footnote{https://mesonet.agron.iastate.edu/request/asos/1min.phtml}. The data consists of 18 weather stations (the \textit{Entities}) distributed in the central US, which provide measurements every minute.  The measurements we considered are wind strength, wind direction, temperature, air pressure, dew point and visibility coefficient (the \textit{Attributes}).

In the analysis we used data from 5 months from April 2008 to August 2008.
The original database consists of 18 tables one for each station.

% We still use three tensors, but with slightly different semantics.
The event tensor is now a sensor tensor with
quadruples  \textit{(Station, measurement, SensorType, Time; Value)}, where \textit{Value} is the sensor measurement for sensor \textit{SensorType} at station \textit{Station} at time \textit{Time}.  The KG-tensor is a long-term memory and   maintains a track record  of sensor measurement history.

% , i.e.,  $x_{s, p, o, t, :} = z_{s, p, o, t-T},  \ldots, z_{s, p, o, t}$.

 As the dataset contains missing values we only considered the periods in which the data is complete. This results in a total of 130442 time steps for our dataset.  In order to capture important patterns in the data and to reduce noise, we applied moving average smoothing using a Hanning window of 21 time steps.
We split the data into train-, validation- and test set. The first four months of the dataset where used for training, and the last month as test set. 5 \% of the training data where used for validation.

We considered three different prediction  models with Gaussian likelihood functions,   each with different latent representations at the input. The first model  (Pred1) is
\[
\theta^{predict}_{s, p, o, t} =
f^{predict}_{p, o}
( {\mathbf{a}}_{e_{s}},
{\mathbf{a}}_{e_{s, t-1}},
{\mathbf{a}}_{e_{s, t-2}}, \ldots,
{\mathbf{a}}_{e_{s, t-T}} )
\]
where
${\mathbf{a}}_{e_{s, t}}$
stands for all measurements of station $e_s$ at time $t$ and ${\mathbf{a}}_{e_{s, t-1}},
{\mathbf{a}}_{e_{s, t-2}}, \ldots,
{\mathbf{a}}_{e_{s, t-T}}$  can be considered a short term memory.
${\mathbf{a}}_{e_{s, t-1}}$
represents all measurements for station $s$ between $t-T-1$ and $t-1$, i.e., and can represent complex sensor patterns over a longer period in time. Since measurements take on real values, a Gaussian likelihood model was used.

The second model (Pred2) is
\[
\theta^{predict}_{s, p, o, t} =
f^{predict}_{p, o}
(
{\mathbf{a}}_{e_{s, t-1}},
 \ldots,
{\mathbf{a}}_{e_{s, t-T}},
{\mathbf{a}}_{e_{t-1}},
 \ldots,
{\mathbf{a}}_{e_{t-T}}
) .
\]
Here,  ${\mathbf{a}}_{e_t} $  stands the latent representation of  all measurements in the complete network at time $t$.

And finally the third model (Pred3) combines the first two models and uses the combined sets of inputs.
 The architecture of Pred3 is shown in Figure~\ref{fig:archs3}.

In our experiments we considered the task of predicting 20 time steps into the future. All three models performed best with $T  = 10$  and the rank of the latent representations being 20.
Table \ref{table:table_sensor_results} summarizes the results of the three prediction models together with three baseline models. The most basic baseline is to use the last observed value of each time series  as a prediction. More enhanced baseline models are  linear regression and feedforward neural networks  using the previous history
$z_{s, :, :, t-1}, z_{s, :, :, t-2}, \ldots, z_{s, :, :, t-T}$
of all time series of a station $s$ as input.
The experiments show that all three prediction models outperform the baselines. Pred1, which adds the personalization term for each sensor shows the best results. Pred2 performs only slightly better than the feedforward neural network. However, we assume that in sensor networks with a stronger cross correlation between the sensors, this model might prove its strength. Finally, the result of Pred3 shows that the combination of the multiple latent representations is too complex and does not outperform Pred1.

\section{Conclusions and Extensions}
\label{sec:concl}

We have introduced an approach for modeling the temporal evolution of knowledge graphs and for the evolution of associated events and signals. We have demonstrated experimentally that  models using latent representations  perform well in these high-dimensional and highly sparse dynamic domains in a clinical application, a recommendation engine and a sensor network application. The clinical application is explored further in a funded project~\cite{Tresp2013,Crist2015}.
As part of future work we plan to test our approach in general streaming frameworks which often contain a context model, an event model and a sensor model, nicely fitting into our framework. In~\cite{tresp2015learning} we are exploring links between the presented approach and cognitive memory functions.

In general, we assumed a unique representation for an entity, for example we assume that ${\mathbf{a}}_{e_{s}}$ is the same in the prediction model and the semantic model.
Sometimes it makes sense to relax that assumption and only assume some form of a coupling.  \cite{larochelle2008zero,bengio2013representation,bengio2012deep} contain extensive discussions on the transfer of latent representations.

% \small
\bibliographystyle{plain}
\bibliography{EventKG}

\section*{Appendix: Cost Functions}

We consider cost functions for the KG tensor, the event tensor and the prediction model. The tilde notation $\underline{\tilde{X}} $ indicates subsets which correspond to  the facts known in training. If only positive  facts with $\textit{Value }= \textit{True}$ are known, as often the case in KGs,  negative facts can  be generated using,  e.g., local closed world assumptions~\cite{nickel2015}. We use negative log-likelihood cost terms. 
 For a Bernoulli likelihood,
$- \log P(x| \theta) = \log [1 + \exp\{(1-2x)\theta\}]$ (cross-entropy) and for a Gaussian likelihood
$- \log P(x| \theta) = const +  \frac{1}{2 \sigma^2}(x-\theta)^2$. We use regularization as described in Equation~\ref{eq:cost}.

We describe the cost function in terms  of the latent representations $A$ and the $M$-mappings. $W$ stands for the parameters in the functional mapping.
\subsubsection*{KG} The cost term for the semantic KG model is
\[
\textrm{cost}^{\textit{KG}} = - \sum_{ x_{s, p, o} \in \underline{\tilde{X}} }  \log P(x_{s, p, o} |
\theta^{\textit{KG}}_{s, p, o} (A, M, W))
\]
\subsubsection*{Events}
\[
\textrm{cost}^{\textit{event}} =
- \sum_{z_{s, p, o, t} \in \underline{{\tilde Z}} }  \log P(z_{s, p, o, t} |
\theta^{\textit{event}}_{s, p, o, t} (A, M, W))
\]
%\subsection*{Event Buffer Memory}
%
%\[
%\textrm{cost}^{\textit{eventBuffer}} =
%- \sum_s  \sum_{y^s_{p, o, t} \in \underline{\tilde{Y}}^s }  \log P(y^s_{p, o, t} |
%\theta^{\textit{eventBuffer}, s}_{ p, o, t} (A, M, W))
%\]
%%
%\subsubsection*{Sensory  Memory}
%\[
%\textrm{cost}^{\textit{sensory}} =
%-  \sum_s  \sum_{  u^s_{q, \gamma, t} \in \underline{\tilde{U}}^s }  \log P(u^s_{q, \gamma, t} |
%\theta^{\textit{sensory}, s}_{q,  \gamma, t} (A, M, W))
%\]
\subsubsection*{Prediction  Model}
The cost function for the prediction model is
\[
\textrm{cost}^{\textit{predict}} =
- \sum_{z_{s, p, o, t} \in \underline{{\tilde Z}}}  \log P(z_{s, p, o, t} | \theta^{\textit{predict}}_{s, p, o, t}
(A, M, W))
\]

\end{document}